\pdfoutput=1
\documentclass[11pt]{article}

\usepackage[preprint]{acl}

\usepackage{times}
\usepackage{latexsym}

\usepackage[T1]{fontenc}
\usepackage[utf8]{inputenc}

\usepackage{microtype}

\usepackage{inconsolata}

\usepackage{graphicx}
\usepackage{enumerate}
\usepackage{multicol}
\usepackage{multirow}
\usepackage{amsmath}
\usepackage{fontawesome}

\title{Memory Reviving, Continuing Learning and Beyond: Evaluation of Pre-trained Encoders and Decoders for Multimodal Machine Translation}

\author{
  Zhuang Yu\textsuperscript{1},
  Shiliang Sun\textsuperscript{1}\thanks{Corresponding author.},
  Jing Zhao\textsuperscript{2},
  Tengfei Song\textsuperscript{3},
  Hao Yang\textsuperscript{3} \\
  \textsuperscript{1}Department of Automation, Shanghai Jiao Tong University \\
  \textsuperscript{2}School of Computer Science and Technology, East China Normal University \\
  \textsuperscript{3}2012 Labs, Huawei Technologies CO., LTD \\
  \texttt{yyyyzzzz@sjtu.edu.cn}, \texttt{shiliangsun@gmail.com}, \texttt{jzhao@cs.ecnu.edu.cn},\\
  \texttt{\{songtengfei2, yanghao30\}@huawei.com}
}

\begin{document}
\maketitle
\begin{abstract}
Multimodal Machine Translation (MMT) aims to improve translation quality by leveraging auxiliary modalities such as images alongside textual input. While recent advances in large-scale pre-trained language and vision models have significantly benefited unimodal natural language processing tasks, their effectiveness and role in MMT remain underexplored. In this work, we conduct a systematic study on the impact of pre-trained encoders and decoders in multimodal translation models. Specifically, we analyze how different training strategies, from training from scratch to using pre-trained and partially frozen components, affect translation performance under a unified MMT framework. Experiments are carried out on the Multi30K and CoMMuTE dataset across English–German and English–French translation tasks. Our results reveal that pre-training plays a crucial yet asymmetrical role in multimodal settings: pre-trained decoders consistently yield more fluent and accurate outputs, while pre-trained encoders show varied effects depending on the quality of visual-text alignment. Furthermore, we provide insights into the interplay between modality fusion and pre-trained components, offering guidance for future architecture design in multimodal translation systems.
\end{abstract}

\section{Introduction}
With the rapid development of deep learning and neural machine translation (NMT), end-to-end translation systems based on the Transformer architecture have achieved remarkable success across various language pairs \cite{vaswani2017attention, ott-etal-2018-scaling}. However, conventional NMT systems rely solely on source language text for modeling, which limits their ability to handle context-dependent translation challenges such as lexical ambiguity and coreference resolution. To address these issues, researchers have begun exploring the integration of multimodal information such as images and audio—into translation models, leading to the emergence of Multimodal Machine Translation (MMT) \cite{elliott-etal-2016-multi30k, specia-etal-2016-shared, calixto-etal-2017-doubly, caglayan-etal-2019-probing, caglayan-etal-2021-cross}. 

The core idea of MMT is to leverage external modalities, such as images, to provide additional semantic cues that can help resolve ambiguities inherent in purely text-based translation. In typical MMT scenarios, the model receives both the source language sentence and an associated image, and jointly encodes both modalities to produce more accurate and context-aware translations \cite{liu2020exploring}. While existing studies have shown that visual information can enhance translation performance in certain cases \cite{caglayan-etal-2019-probing}, the overall improvements remain inconsistent, and there is a lack of standardized evaluation frameworks.

\begin{figure}
    \centering
    \includegraphics[width=\linewidth]{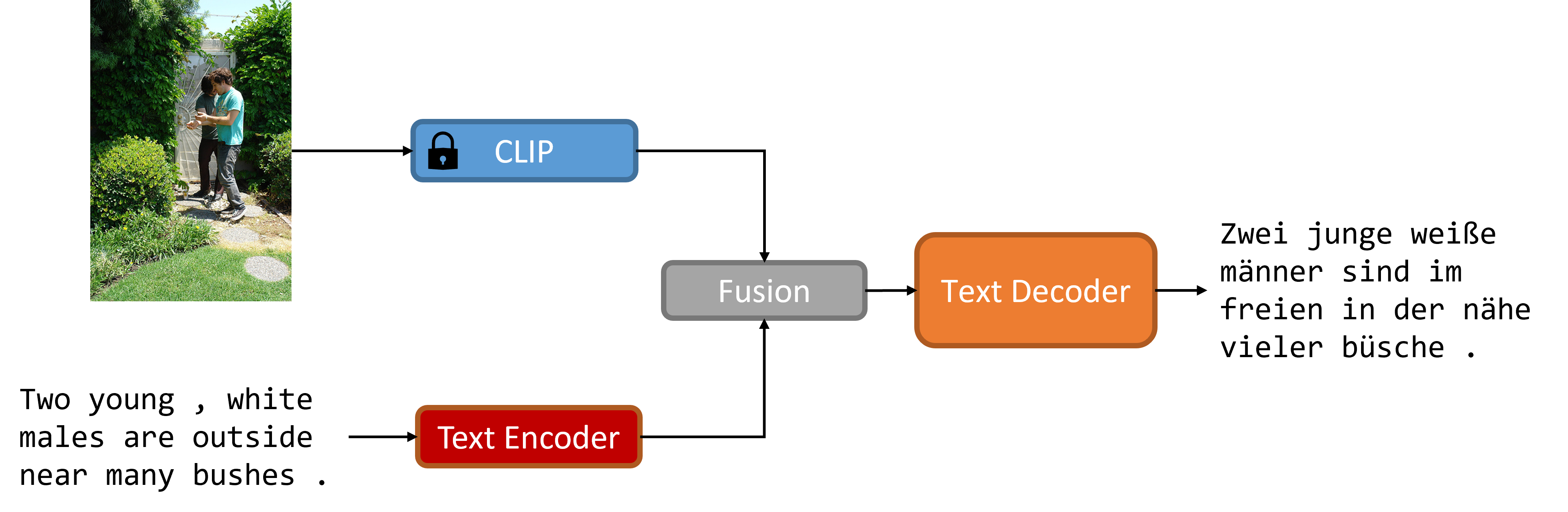}
    \caption{Our unified MMT baseline. For the visual encoder, we use CLIP to extract image features. For the text encoder and decoder, we will explore the impact of the pre-trained components on the entire baseline.}
    \label{fig:pretrained model}
\end{figure}

Meanwhile, the emergence of large-scale pre-trained models has greatly advanced both natural language processing and multimodal learning. Models such as BERT \cite{devlin-etal-2019-bert} and mBART \cite{liu-etal-2020-multilingual-denoising} have demonstrated strong cross-lingual transfer capabilities, particularly in multilingual understanding and generation tasks such as machine translation, while vision-language models like CLIP \cite{DBLP:journals/corr/abs-2103-00020} and BLIP \cite{DBLP:journals/corr/abs-2201-12086} provide high quality image semantic representations. However, in the context of MMT, the integration of pre-trained models is often empirically driven: some studies utilize pre-trained language models to initialize the encoder, while others adopt visual features as auxiliary inputs. However, a systematic analysis and comparative evaluation of how different pre-trained components function within MMT systems remains largely absent.

This work aims to fill that gap by focusing on the role and impact of pre-trained encoders and decoders in MMT. Within a unified MMT framework, as shown in Figure \ref{fig:pretrained model}, we systematically investigate the effects of various training strategies, including training from scratch, freezing, and fine-tuning, across both the encoder and decoder components. We conduct comparative experiments on the English-German and English-French translation tasks using the Multi30K and CoMMuTE datasets. Our results reveal a clear asymmetry in the contributions of pre-trained modules: decoder pre-training provides consistently stable gains in generation quality, while encoder pre-training is more sensitive to the alignment quality between visual and textual semantics. Our findings offer new empirical insights to guide the design and training of future multimodal translation systems.

Our main contributions are summarized as follows:
\begin{itemize}
    \item \textbf{Comprehensive evaluation of pretrained components in MMT.} We provide a systematic analysis of how pre-trained encoders and decoders contribute to MMT, under various initialization and training strategies (e.g., frozen, fine-tuned, or trained from scratch).
    \item \textbf{Empirical insights into asymmetric effects of pre-training.} Our experiments reveal a modality-dependent asymmetry: pre-trained decoders offer stable improvements in generation quality, while the benefit of pretrained encoders strongly depends on the degree of visual-textual alignment.
    \item \textbf{Unified framework and established strong MMT baselines.} Our framework supports flexible integration of pre-trained modules with different training strategies, and we benchmark English–German and English–French multimodal translation on Multi30K and CoMMuTE, offering reproducible baselines for future research.
\end{itemize}

\section{Related Work}
\subsection{Pre-trained Language Models for Machine Translation}
In recent years, pre-trained language models such as BERT \cite{devlin-etal-2019-bert} and mBART \cite{liu-etal-2020-multilingual-denoising} have been widely adopted in Neural Machine Translation (NMT), significantly enhancing models' language understanding and cross-lingual generalization capabilities. Particularly in multilingual settings, models like mBART and T5 \cite{DBLP:journals/corr/abs-1910-10683}, trained on large scale cross-lingual corpora, have demonstrated strong performance in multilingual representation learning, becoming foundational components for building translation systems across diverse language pairs.

With the rise of large language models (LLMs) such as GPT \cite{radford2018improving, radford2019language, DBLP:journals/corr/abs-2005-14165, openai2024gpt4technicalreport}, PaLM \cite{chowdhery2022palmscalinglanguagemodeling, anil2023palm2technicalreport} and XLM-R \cite{conneau-etal-2020-unsupervised}, machine translation has further benefited from their powerful knowledge transfer abilities, especially in zero-shot and low-resource scenarios. Recent efforts have also explored incorporating LLMs into MMT; for instance, \citet{hendy2023good} evaluated ChatGPT and GPT-3.5 in 18 translation directions. However, the impact mechanisms of different training strategies within MMT systems remain unclear. This work aims to address this gap by systematically analyzing the synergy and divergence of various pre-trained components in MMT.

\subsection{Vision-Language Alignment and Visual Semantics in MMT}
In MMT, the visual modality can provide intuitive semantic cues that are closely related to the source text, offering significant advantages in handling concrete descriptions, coreference resolution, and scene understanding \cite{elliott-etal-2016-multi30k}. Early studies typically adopted an “image-assisted” approach, integrating image features—extracted via Convolutional Neural Networks (CNNs) with textual representations \cite{calixto-liu-2017-incorporating}, or modeling the two modalities through dual stream architectures followed by cross-modal attention mechanisms \cite{caglayan-etal-2016-multimodality}. With the emergence of cross-modal pre-trained models such as CLIP \cite{DBLP:journals/corr/abs-2103-00020} and ALBEF \cite{DBLP:journals/corr/abs-2107-07651}, the quality of visual semantic representations has been greatly improved, enabling more accurate vision-language alignment in MMT tasks. Subsequently, \citet{li-etal-2022-vision} introduced selective attention to further refine how models interpret visual information, and GRAM architecture \cite{vijayan-etal-2024-adding} proposed an alternative translation framework that explicitly integrates visual grounding mechanisms.

Nevertheless, incorporating visual information into MMT still faces several open challenges \cite{vijayan-etal-2024-adding, wu-etal-2021-good}. On the one hand, image features may include redundant or irrelevant details, potentially introducing noise into the language modeling process. On the other hand, the significance of visual semantics varies greatly across translation samples, and there is still no unified paradigm for dynamically regulating the degree of visual involvement. As a result, evaluating the utility and necessity of visual features, as well as exploring more effective fusion mechanisms with textual semantics, remains a key research direction in the field of MMT.

\section{Experimental Setup}
In this section, we will introduce the datasets used in our experiments, the various pre-trained components used, and the corresponding evaluation metrics.

\subsection{Datasets}
We conduct experiments on two MMT benchmarks: Multi30k \cite{elliott-etal-2016-multi30k} and CoMMuTE \cite{futeral-etal-2023-tackling}:

\textbf{Multi30k} mainly contains two translation directions: English-German and English-French, including 31014 images with English captions and German and French translations. The training set and validation set contain 29,000 and 1,014 instances, respectively. We reported the results on the Test2016, Test2017 and MSCOCO test sets, which contain 1000, 1000 and 461 instances, respectively.

\textbf{CoMMuTE} is a contrastive multilingual multimodal translation evaluation dataset, which covers three translation directions: English-French, English-German, and English-Czech. It consists of 155 lexically ambiguous English sentences, each with two translations and two possible meanings, and two pictures to determine which translation is correct.

\subsection{Pre-trained Encoders and Decoders}
We systematically explore the impact of different pre-training strategies on the encoder and decoder, covering a variety of combinations from traditional Transformers to large pre-trained language models.

For the image encoder, we use the \textbf{CLIP-ViT} model \footnote{\url{https://huggingface.co/openai/clip-vit-large-patch14}}, which is widely used in image-text alignment tasks, to extract image features as visual input. Since CLIP has good migration capabilities on large-scale image-text paired data, we choose to freeze all its parameters in this experiment to avoid the unstable effect of image encoding on the training process.

For the text encoders and decoders, we divide the models into two categories according to their structural characteristics:

(i) \textbf{Encoder-Decoder structural models}, including basic Transformer (three sizes: Small / Base / Large) \cite{vaswani2017attention}, T5-small \cite{DBLP:journals/corr/abs-1910-10683} and mBART-large \cite{liu-etal-2020-multilingual-denoising}. This type of structure has good bidirectional understanding (Encoder) and autoregressive generation (Decoder) capabilities, and is often used in classic machine translation scenarios.

(ii) \textbf{Decoder-only structural models}, including lightweight large language models Qwen2.5-0.5B \cite{qwen2025qwen25technicalreport} and LLaMA3.2-1B \cite{grattafiori2024llama3herdmodels}. As representative general LLMs, they directly rely on context modeling capabilities for translation generation in the decoding stage and visual information is often incorporated as textual prompts, directly participating in context modeling and memory reviving. 

Table \ref{tab:model architecture} summarizes the structural parameters of various encoders and decoders:
\begin{table}[t]
    \centering
    \resizebox{1.0\linewidth}{!}{
    \begin{tabular}{c|c|c|c}
        \hline
        \textbf{Model Name} & \textbf{Layers} & \textbf{Hidden Dim} & \textbf{Attention Heads} \\
        \hline
        Transformer-Small  & 4   & 256   & 4  \\
        Transformer-Base   & 6  & 512   & 8 \\
        Transformer-Large  & 8  & 1024  & 16 \\
        T5-Small           & 6   & 512   & 8  \\
        mBART-Large        & 12  & 1024  & 16 \\
        Qwen2.5-0.5B       & 16  & 2048  & 16 \\
        LLaMA3.2-1B        & 24  & 3072  & 24 \\ 
        \hline
    \end{tabular}
    }
    \caption{Architecture details of text encoders and decoders used in our experiments.}
    \label{tab:model architecture}
\end{table}

\begin{table*}[t]
    \centering
    \resizebox{1.0\linewidth}{!}{
    \begin{tabular}{c|c c c|c c c|c c c|c c c}
        \hline
        \multicolumn{13}{c}{\textbf{En-De}} \\
        \hline
        \multirow{2}{*}{\textbf{Models}} & \multicolumn{3}{|c|}{Test2016} & \multicolumn{3}{|c|}{Test2017} & \multicolumn{3}{|c|}{MSCOCO} & \multicolumn{3}{|c}{CoMMuTE} \\ \cline{2-13}
         & BLEU & METEOR & COMET & BLEU & METEOR & COMET & BLEU & METEOR & COMET & BLEU & METEOR & COMET \\ \hline
        Transformer-Small & 43.2 & 69.9 & 34.2 & 35.8 & 63.5 & 32.0 & 31.7 & 58.3 & 26.3 & 5.3 & 19.8 & 10.0 \\
        Transformer-Base & 43.6 & 70.2 & 36.5 & 35.8 & 63.6 & 32.7 & 32.4 & 59.2 & 27.2 & 5.4 & 20.0 & 10.6 \\
        Transformer-Large & 44.5 & 70.8 & 38.1 & 36.7 & 64.0 & 33.4 & 33.5 & 59.8 & 29.5 & 5.8 & 21.2 & 11.1 \\
        T5-Small & 92.3 & 96.5 & 81.6 & 81.5 & 96.1 & 78.1 & 91.2 & 95.8 & 80.4 & 15.4 & 44.3 & 29.7 \\
        mBART-Large & 98.7 & 99.2 & 91.4 & 98.9 & \textbf{99.3} & 91.2 & 99.1 & 99.4 & 92.6 & 27.8 & 48.7 & 37.1 \\
        Qwen2.5-0.5B & 98.7 & \textbf{99.3} & \textbf{93.1} & 98.8 & \textbf{99.3} & 92.8 & 99.1 & \textbf{99.5} & \textbf{93.3} & 30.0 & \textbf{51.8} & 40.7 \\
        LLaMA3.2-1B & \textbf{98.9} & 99.0 & 92.9 & \textbf{99.0} & 99.2 & \textbf{93.4} & \textbf{99.2} & 99.4 & 92.7 & \textbf{30.2} & 51.5 & \textbf{43.4} \\
        \hline
        \multicolumn{13}{c}{\textbf{En-Fr}} \\
        \hline
        Transformer-Small & 62.8 & 81.7 & 71.0 & 55.3 & 76.7 & 64.7 & 46.2 & 71.0 & 50.3 & 7.8 & 22.9 & 11.5 \\
        Transformer-Base & 63.4 & 82.1 & 73.2 & 55.9 & 77.0 & 66.1 & 46.7 & 71.6 & 51.5 & 8.1 & 23.4 & 11.8 \\
        Transformer-Large & 64.3 & 82.4 & 77.2 & 57.0 & 77.6 & 69.2 & 47.4 & 72.0 & 54.7 & 8.9 & 25.0 & 12.4 \\
        T5-Small & 95.5 & 98.1 & 83.4 & 84.4 & 98.3 & 80.8 & 94.2 & 97.8 & 82.3 & 50.1 & 66.3 & 47.2 \\
        mBART-Large & 99.0 & 99.1 & \textbf{93.4} & \textbf{99.0} & 99.1 & 93.1 & 99.0 & \textbf{99.5} & 91.9 & 60.1 & 79.0 & 58.1 \\
        Qwen2.5-0.5B & 99.1 & \textbf{99.2} & 92.5 & 98.9 & \textbf{99.2} & \textbf{94.2} & \textbf{99.1} & 99.4 & \textbf{93.3} & 62.2 & 78.4 & \textbf{61.2} \\
        LLaMA3.2-1B & \textbf{99.2} & 99.1 & 92.8 & \textbf{99.0} & \textbf{99.2} & 93.9 & \textbf{99.1} & 99.3 & 93.1 & \textbf{62.5} & \textbf{79.2} & 60.8 \\
        \hline
    \end{tabular}
    }
    \caption{MMT performance (BLEU, METEOR, and COMET) of various models on Multi30k and CoMMuTE datasets for En-De and En-Fr directions. In the unified MMT baseline mentioned above, we perform full parameter fine-tuning on the encoder or decoder of the pre-trained model, rather than training and fine-tuning the entire model.}
    \label{tab:overall translation}
\end{table*}

\subsection{Evaluation Metrics}
In order to comprehensively evaluate the performance of different training strategies in MMT tasks, this paper adopts three mainstream machine translation evaluation indicators: BLEU \cite{papineni-etal-2002-bleu}, METEOR \cite{banerjee-lavie-2005-meteor} and COMET \cite{rei-etal-2020-comet}, which measure the accuracy and readability of the translation output from the two dimensions of surface matching and semantic quality. 

\textbf{BLEU (Bilingual Evaluation Understudy)} is a precision metric that calculates the degree of n-gram overlap between the candidate and reference translations. The BLEU score is defined as:

\begin{equation}
\text{BLEU} = BP \cdot \exp\left( \sum_{n=1}^{N} w_n \log p_n \right)
\end{equation}
where $p_n$ is the precision of n-gram matches, $w_n$ are weights (typically uniform), and $BP$ is the brevity penalty to penalize short translations.

\textbf{METEOR (Metric for Evaluation of Translation with Explicit ORdering)} goes beyond surface matching by incorporating stemming, synonymy, and paraphrase matches. It computes an F-score-based harmonic mean of precision and recall, adjusted by a penalty for incorrect word order:

\begin{equation}
\text{METEOR} = F_{\text{mean}} \cdot (1 - Penalty)
\end{equation}
where $F_{\text{mean}} = \frac{10 \cdot P \cdot R}{R + 9P}$, and the Penalty is based on the number of chunks (contiguous matches).

\textbf{COMET (Crosslingual Optimized Metric for Evaluation of Translation)} is a neural-based metric built on pretrained language models. It evaluates semantic similarity by comparing the embeddings of the source sentence, reference, and candidate translations using a regression model fine-tuned on human judgments. Though COMET does not have a simple closed-form equation like BLEU or METEOR, it can be abstractly expressed as:

\begin{equation}
\text{COMET}(x, y, \hat{y}) = f_{\theta}\left( \text{Enc}(x), \text{Enc}(y), \text{Enc}(\hat{y}) \right)
\end{equation}
where $x$, $y$ and $\hat{y}$ are the source, reference, and candidate translations respectively, $\text{Enc}$ denotes the encoder (e.g., XLM-R), and $f_{\theta}$ is a learned scoring function.

\begin{table*}[t]
    \centering
    \resizebox{1.0\linewidth}{!}{
    \begin{tabular}{c|c c c|c c c|c c c|c c c}
        \hline
        \multicolumn{13}{c}{\textbf{En-De}} \\
        \hline
        \multirow{2}{*}{\textbf{Models}} & \multicolumn{3}{c|}{Test2016} & \multicolumn{3}{c|}{Test2017} & \multicolumn{3}{c|}{MSCOCO} & \multicolumn{3}{c}{CoMMuTE} \\
        \cline{2-13}
        & BLEU & METEOR & COMET & BLEU & METEOR & COMET & BLEU & METEOR & COMET & BLEU & METEOR & COMET \\
        \hline
        T5-Encoder + Trans-Dec & 47.3 & 71.9 & 58.5 & 38.5 & 65.7 & 51.0 & 35.0 & 60.2 & 46.8 & 6.1 & 25.5 & 13.2 \\
        mBART-Encoder + Trans-Dec & 48.1 & 72.5 & 59.6 & 39.2 & 66.4 & 52.1 & 35.6 & 61.1 & 47.9 & 6.3 & 26.1 & 13.0 \\
        Qwen-Embed + Trans-Dec & \textbf{49.8} & \textbf{73.0} & \textbf{61.0} & \textbf{41.0} & 67.0 & 53.2 & \textbf{37.0} & \textbf{62.0} & \textbf{49.2} & 6.9 & 27.4 & 15.4 \\
        LLaMA-Embed + Trans-Dec & 49.2 & 72.8 & 60.4 & 40.6 & \textbf{67.3} & \textbf{53.8} & 36.8 & 61.8 & 48.9 & \textbf{7.5} & \textbf{27.9} & \textbf{16.8} \\
        \hline
        Trans-Enc + T5-Decoder & 46.5 & 70.8 & 56.7 & 38.0 & 64.9 & 49.5 & 34.4 & 59.3 & 45.0 & 6.5 & 24.1 & 13.0 \\
        Trans-Enc + mBART-Decoder & 47.1 & 71.5 & 57.9 & 38.8 & 65.5 & 50.7 & 34.9 & 60.1 & 46.1 & 6.3 & 24.7 & 13.7 \\
        Trans-Enc + Qwen & \textbf{50.3} & \textbf{73.4} & \textbf{61.6} & \textbf{41.5} & \textbf{67.6} & \textbf{54.4} & \textbf{37.5} & \textbf{62.3} & \textbf{49.6} & 8.0 & 26.5 & 15.5 \\
        Trans-Enc + LLaMA & 49.9 & 73.1 & 60.9 & 41.2 & 67.4 & 54.0 & 37.3 & 62.1 & 49.3 & \textbf{8.7} & \textbf{27.2} & \textbf{17.0} \\
        \hline
        \multicolumn{13}{c}{\textbf{En-Fr}} \\
        \hline
        T5-Encoder + Trans-Dec & 63.3 & 84.0 & 64.5 & 56.7 & 77.1 & 60.2 & 47.0 & 72.1 & 54.3 & 11.0 & 31.5 & 21.2 \\
        mBART-Encoder + Trans-Dec & 62.9 & 83.4 & 64.2 & 57.3 & 76.5 & 59.0 & 46.8 & 71.6 & 54.2 & 12.3 & 32.1 & 22.0 \\
        Qwen-Embed + Trans-Dec & 64.2 & 85.1 & 65.6 & 57.6 & 77.3 & 61.5 & 47.8 & \textbf{72.8} & 54.8 & 14.1 & 39.7 & 25.4 \\
        LLaMA-Embed + Trans-Dec & \textbf{64.9} & \textbf{85.9} & \textbf{66.2} & \textbf{58.3} & \textbf{78.0} & \textbf{62.2} & \textbf{48.6} & 72.5 & \textbf{55.5} & \textbf{14.5} & \textbf{40.2} & \textbf{26.8} \\
        \hline
        Trans-Enc + T5-Decoder & 63.6 & 84.6 & 65.9 & 57.0 & 76.6 & 59.1 & 47.3 & 70.6 & 54.2 & 12.5 & 35.1 & 21.0 \\
        Trans-Enc + mBART-Decoder & 63.2 & 84.9 & 65.7 & 57.5 & 76.9 & 59.8 & 47.8 & 69.9 & 55.1 & 13.3 & 38.7 & 21.7 \\
        Trans-Enc + Qwen & 65.6 & \textbf{86.3} & \textbf{66.0} & 58.1 & 78.5 & 63.0 & \textbf{48.2} & 73.2 & 55.5 & \textbf{16.0} & \textbf{42.5} & \textbf{28.5} \\
        Trans-Enc + LLaMA & \textbf{66.3} & 86.2 & 65.7 & \textbf{58.8} & \textbf{79.2} & \textbf{63.8} & 48.0 & \textbf{74.0} & \textbf{56.1} & 15.7 & 41.2 & 26.0 \\
        \hline
    \end{tabular}
    }
    \caption{MMT performance with different pre-trained encoders and decoders on the En-De and En-Fr translation task. Respectively, Trans-Enc and Trans-Dec represent the standard Transformer encoder and decoder.}
    \label{tab:pre-trained encoder and decoder}
\end{table*}

\section{Results and Analysis}
\subsection{Overall Translation Performance}
As shown in Table \ref{tab:overall translation}, in order to evaluate the performance of various models on the Multi30k and CoMMuTE datasets, we train them separately on two 4090 GPUs until convergence. From the results, we observe several notable trends regarding the impact of training and model architecture on MMT.

First, pre-trained language models significantly outperform vanilla Transformer models across all evaluation metrics. This performance gap is particularly pronounced on the Multi30k test sets, suggesting that pre-trained models possess stronger semantic representation and generalization capabilities, likely inherited from their large-scale multilingual training.

Second, decoder-only large language models such as Qwen and LLaMA achieve particularly strong results across evaluation sets, especially on the open-domain CoMMuTE dataset. These models benefit from extensive pre-training on diverse textual data, enabling them to generalize well even in multimodal translation tasks, despite not being specifically trained for translation. Their flexibility and broad knowledge coverage allow them to better handle ambiguous or less constrained input scenarios.

Third, while encoder-decoder models like mBART still perform competitively on more structured datasets like Multi30k, their performance tends to degrade in less conventional or domain-shifted settings. This indicates that although such architectures benefit from task-specific design, they may lack the robustness and adaptability demonstrated by decoder-only LLMs.

Finally, scaling up traditional Transformer models (from Small to Large) yields marginal improvements, but still falls significantly short compared to pretrained counterparts. This underscores that increasing model capacity alone is insufficient to match the benefits provided by comprehensive language pre-training.

\subsection{Memory Reviving and Continuing Learning}
\begin{figure}
    \centering
    \includegraphics[width=1.0\linewidth]{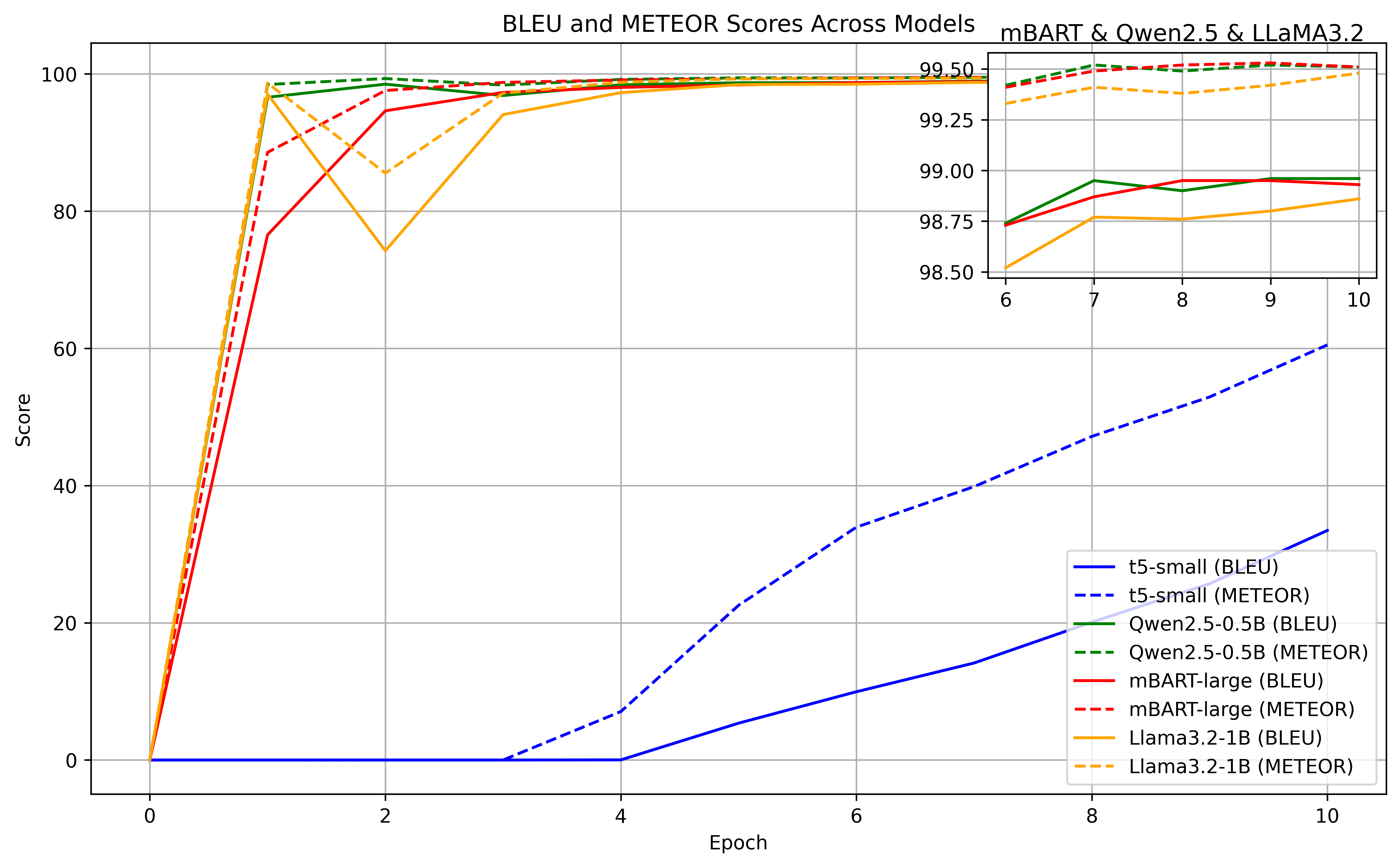}
    \caption{The trend of BLEU and METEOR scores of the pre-trained model from scratch with epoch on Multi30k.}
    \label{fig:metrics comparison}
\end{figure}

\begin{table*}[t]
    \centering
    \resizebox{1.0\linewidth}{!}{
    \begin{tabular}{c|c|c c c|c c c|c c c|c c c}
        \hline
        \multicolumn{14}{c}{\textbf{En-De}} \\
        \hline
        \multirow{2}{*}{\textbf{Models}} & \multirow{2}{*}{Visual Modality} & \multicolumn{3}{c|}{Test2016} & \multicolumn{3}{c|}{Test2017} & \multicolumn{3}{c|}{MSCOCO} & \multicolumn{3}{c}{CoMMuTE}\\
        \cline{3-14}
        & & BLEU & METEOR & COMET & BLEU & METEOR & COMET & BLEU & METEOR & COMET & BLEU & METEOR & COMET \\
        \hline 
        \multirow{2}{*}{T5-Small} & \faCheck & 92.3 & 96.5 & 81.6 & 81.5 & 96.1 & 78.1 & 91.2 & 95.8 & 80.4 & 15.4 & 44.3 & 29.7 \\
        & \faTimes & 93.0($\uparrow$ 0.7) & 97.0($\uparrow$ 0.5) & 82.0($\uparrow$ 0.4) & 82.2($\uparrow$ 0.7) & 91.6($\uparrow$ 0.5) & 78.7($\uparrow$ 0.6) & 91.8($\uparrow$ 0.6) & 96.5($\uparrow$ 0.7) & 81.0($\uparrow$ 0.6) & 17.0($\uparrow$ 1.6) & 45.8($\uparrow$ 1.5) & 30.5($\uparrow$ 0.8) \\ \hline
        \multirow{2}{*}{mBART-Large} & \faCheck & 98.7 & 99.2 & 91.4 & 98.9 & 99.3 & 91.2 & 99.1 & 99.4 & 92.6 & 27.8 & 48.7 & 37.1 \\
        & \faTimes & 98.8($\uparrow$ 0.1) & 99.3($\uparrow$ 0.1) & 91.6($\uparrow$ 0.2) & 99.0($\uparrow$ 0.1) & \textbf{99.5}($\uparrow$ 0.2) & 91.3($\uparrow$ 0.1) & \textbf{99.2}($\uparrow$ 0.1) & 99.5($\uparrow$ 0.1) & 93.0($\uparrow$ 0.4) & 28.5($\uparrow$ 0.7) & 49.0($\uparrow$ 0.3) & 37.7($\uparrow$ 0.6) \\ \hline
        \multirow{2}{*}{Qwen2.5-0.5B} & \faCheck & 98.7 & 99.3 & 93.1 & 98.8 & 99.3 & 92.8 & 99.1 & 99.5 & 93.3 & 30.0 & \textbf{51.8} & \textbf{40.7} \\
        & \faTimes & 98.9($\uparrow$ 0.2) & \textbf{99.4}($\uparrow$ 0.1) & \textbf{93.4}($\uparrow$ 0.3) & 99.0($\uparrow$ 0.2) & 99.4($\uparrow$ 0.1) & 93.0($\uparrow$ 0.2) & \textbf{99.2}($\uparrow$ 0.1) & \textbf{99.6}($\uparrow$ 0.1) & \textbf{93.5}($\uparrow$ 0.2) & 29.8($\downarrow$ 0.2) & 51.5($\downarrow$ 0.3) & 40.4($\downarrow$ 0.3) \\ \hline
        \multirow{2}{*}{LLaMA3.2-1B} & \faCheck & 98.9 & 99.0 & 92.9 & 99.0 & 99.2 & 93.4 & 99.2 & 99.4 & 92.7 & \textbf{30.2} & 51.5 & 43.4 \\
        & \faTimes & \textbf{99.0}($\uparrow$ 0.1) & 99.2($\uparrow$ 0.2) & 93.0($\uparrow$ 0.1) & \textbf{99.1}($\uparrow$ 0.1) & 93.4($\uparrow$ 0.2) & \textbf{96.5}($\uparrow$ 0.1) & \textbf{99.2}($\uparrow$ 0.1) & \textbf{99.6}($\uparrow$ 0.2) & 92.9($\uparrow$ 0.2) & 30.0($\downarrow$ 0.2) & 51.4($\downarrow$ 0.1) & 43.1($\downarrow$ 0.3) \\ \hline
        \multicolumn{14}{c}{\textbf{En-Fr}} \\
        \hline
        \multirow{2}{*}{T5-Small} & \faCheck & 95.5 & 98.1 & 83.4 & 84.4 & 98.3 & 80.8 & 94.2 & 97.8 & 82.3 & 50.1 & 66.3 & 47.2 \\
        & \faTimes & 96.0($\uparrow$ 0.5) & 98.5($\uparrow$ 0.4) & 83.9($\uparrow$ 0.5) & 84.7($\uparrow$ 0.3) & 94.6($\uparrow$ 0.2) & 81.0($\uparrow$ 0.2) & 94.6($\uparrow$ 0.4) & 98.0($\uparrow$ 0.2) & 82.7($\uparrow$ 0.4) & 51.0($\uparrow$ 0.9) & 66.8($\uparrow$ 0.5) & 48.0($\uparrow$ 0.8) \\ \hline
        \multirow{2}{*}{mBART-Large} & \faCheck & 99.0 & 99.1 & 93.4 & 99.0 & 99.1 & 93.1 & 99.0 & 99.5 & 91.9 & 60.1 & 79.0 & 58.1 \\
        & \faTimes & 99.3($\uparrow$ 0.3) & 99.2($\uparrow$ 0.1) & \textbf{93.6}($\uparrow$ 0.2) & \textbf{99.1}($\uparrow$ 0.1) & \textbf{99.5}($\uparrow$ 0.4) & 93.3($\uparrow$ 0.2) & 99.2($\uparrow$ 0.2) & \textbf{99.7}($\uparrow$ 0.2) & 92.0($\uparrow$ 0.1) & 60.5($\uparrow$ 0.4) & 79.5($\uparrow$ 0.5) & 58.5($\uparrow$ 0.4) \\ \hline
        \multirow{2}{*}{Qwen2.5-0.5B} & \faCheck & 99.1 & 99.2 & 92.5 & 98.9 & 99.2 & 94.2 & 99.1 & 99.4 & 93.3 & 62.2 & 78.4 & \textbf{61.2} \\
        & \faTimes & 99.4($\uparrow$ 0.3) & \textbf{99.4}($\uparrow$ 0.2) & 93.0($\uparrow$ 0.5) & 99.0($\uparrow$ 0.1) & 99.3($\uparrow$ 0.1) & \textbf{94.5}($\uparrow$ 0.3) & 99.2($\uparrow$ 0.1) & \textbf{99.7}($\uparrow$ 0.3) & \textbf{93.5}($\uparrow$ 0.2) & 61.8($\downarrow$ 0.4) & 77.9($\downarrow$ 0.5) & 60.4($\downarrow$ 0.8) \\ \hline
        \multirow{2}{*}{LLaMA3.2-1B} & \faCheck & 99.2 & 99.1 & 92.8 & 99.0 & 99.2 & 93.9 & 99.1 & 99.3 & 93.1 & \textbf{62.5} & \textbf{79.0} & 60.8 \\
        & \faTimes & \textbf{99.4}($\uparrow$ 0.2) & 99.3($\uparrow$ 0.2) & 93.1($\uparrow$ 0.3) & \textbf{99.1}($\uparrow$ 0.1) & 99.4($\uparrow$ 0.2) & 94.1($\uparrow$ 0.2) & \textbf{99.3}($\uparrow$ 0.2) & 99.4($\uparrow$ 0.1) & 93.3($\uparrow$ 0.2) & 61.6($\downarrow$ 0.9) & 77.9($\downarrow$ 1.1) & 60.2($\downarrow$ 0.6) \\ \hline
    \end{tabular}
    }
    \caption{Evaluation of different pre-trained models with and without visual modality on Multi30k and CoMMuTE test sets. \faTimes indicates that image information is not used in the translation process.}
    \label{fig: visual modality}
\end{table*}

Figure \ref{fig:metrics comparison} illustrates the evolution of BLEU and METEOR scores across training epochs for four models: T5-Small, Qwen2.5-0.5B, mBART-Large, and LLaMA3.2-1B when trained from scratch on Multi30K. This figure provides insight into how well different pre-trained architectures can either memory reviving or engage in continuing learning.

Notably, the large scale pre-trained models: mBART, Qwen, and LLaMA exhibit rapid performance gains during the initial training (epochs 1–2), quickly approaching near-saturated BLEU and METEOR scores. This indicates that these large pre-trained models can effectively revive prior linguistic knowledge and cross-modal alignment capabilities, enabling fast convergence. However, this early stage surge is sometimes accompanied by metric level fluctuations, suggesting some instability as the models adapt to the new multimodal setting and activate their pre-trained memory.

In the later stages of training (epochs 6–10), these models exhibit remarkable consistency, with only marginal variance in both BLEU and METEOR scores. This plateau indicates a high degree of learning stability and resilience to forgetting, affirming the benefits of pre-training when adapting to multimodal translation tasks. In contrast, T5-Small, with its significantly smaller capacity and limited representational prior, shows a steadily rising curve across both metrics. Its performance remains far below that of the larger models throughout training, underscoring its reliance on gradual accumulation of task specific knowledge rather than effective reuse of pre-trained memory.

Nevertheless, despite the early advantages of large pre-trained models, their performance on more complex datasets remains constrained. These results suggest that memory reviving alone is insufficient to fully address the challenges of fine-grained semantic grounding and multimodal alignment. To achieve optimal performance, continuing learning remains essential not only to refine modality interactions, but also to mitigate the gap between pre-training and downstream tasks. Further task-specific training and alignment optimization are thus necessary to fully unlock the potential of large-scale multimodal models.

\subsection{Impact of Pre-trained Encoders and Decoders}
To investigate the effect of different pretraining strategies on multimodal machine translation (MMT), we design two experimental settings: (1) varying the encoder while fixing the decoder as a standard Transformer decoder, and (2) varying the decoder while keeping the encoder as a standard Transformer encoder. We evaluate all models on three Multi30k subsets (Test2016, Test2017, and MSCOCO) and the CoMMuTE dataset, using BLEU, METEOR, and COMET as evaluation metrics.

As shown in Table \ref{tab:pre-trained encoder and decoder}, in the first setting, where we compare different pre-trained encoders, we observe that replacing the vanilla Transformer encoder with pretrained models such as T5 or mBART significantly improves translation performance across all benchmarks. These encoder-decoder models benefit from rich linguistic knowledge acquired during pretraining, enhancing both accuracy (BLEU) and fluency (METEOR). More notably, compared to models such as Qwen and LLaMA, which only use the decoder as a sentence encoder (by extracting its final layer representation), their advantage is significantly less stable on the more diverse and visually complex CoMMuTE dataset. This suggests that the benefit of encoder pre-training depends on how well the visual input is semantically aligned with the source text. When the visual modality is noisy or less correlated, the semantic representations generated by the pre-trained encoder may confuse rather than help translation.

In the second setting, where we fix the encoder and vary the decoder, decoder-only models again outperform traditional pre-trained decoders. While T5 and mBART show moderate improvements over the vanilla Transformer decoder, models like Qwen and LLaMA achieve consistently higher scores across all datasets and metrics. Their autoregressive generation capabilities, combined with deep self-attention mechanisms, enable them to model complex linguistic patterns more effectively. This advantage is particularly evident in the COMET scores on CoMMuTE, highlighting their ability to preserve semantic consistency in challenging multimodal scenarios.

\begin{figure*}
    \centering
    \includegraphics[width=\linewidth]{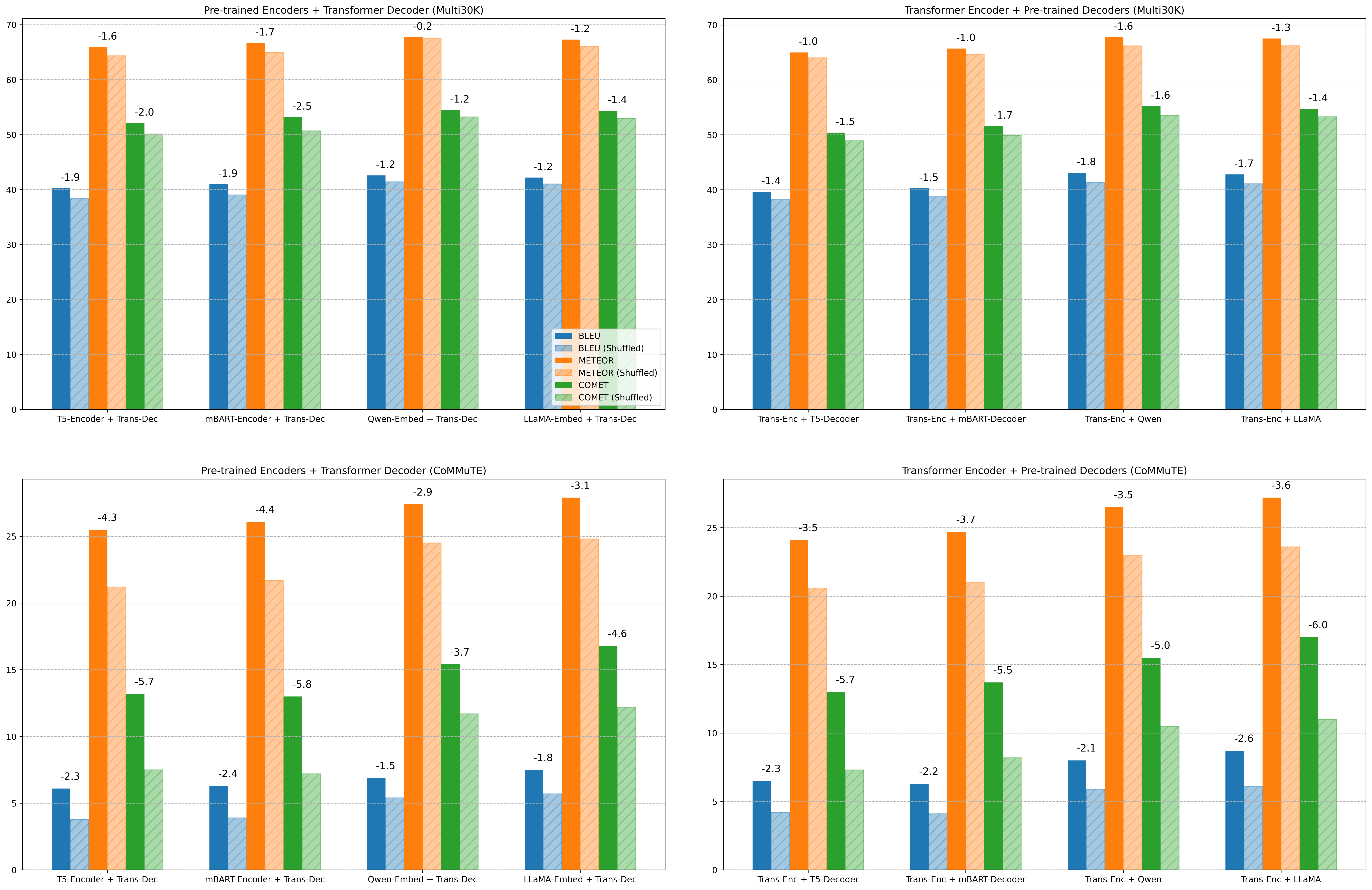}
    \caption{Evaluation of pre-trained encoders and decoders in the En-De direction of the Multi30k and CoMMuTE datasets, comparing the performance of different models in two scenarios: normal alignment and shuffled alignment.}
    \label{fig:final comparison}
\end{figure*}

\subsection{Effect of Visual Modality}
Next, we conduct a systematic comparison across multiple pre-trained models (T5-Small, mBART-Large, Qwen2.5-0.5B, and LLaMA3.2-1B), examining the impact of incorporating visual modality. We evaluate performance using BLEU, METEOR, and COMET scores. As shown in Table \ref{fig: visual modality}: in most cases, adding image information does not improve performance and even leads to degradation across several test sets, suggesting that the visual modality has not yet demonstrated its potential and may even introduce noise.

For encoder-decoder architectures, visual information is typically integrated on the encoder side alongside the source text, aiming to enrich the semantic space of the encoded representations. However, our results show that this integration fails to yield consistent improvements. For example, mBART exhibits slightly lower BLEU, METEOR, and COMET scores when visual inputs are included, both on the Multi30k benchmark and the more challenging CoMMuTE dataset. This suggests that, despite the structural capacity of the encoder to handle multimodal input, image semantics often fail to align properly with textual semantics. As a result, the visual features do not translate into useful contextual signals and may even introduce semantic noise. This issue is especially pronounced when the sentence meaning is already clear and contextually complete, making image content potentially redundant or conflicting, thus misleading the decoder.

In contrast, decoder-only models like Qwen2.5 and LLaMA3.2 exhibit the opposite behavior on the CoMMuTE dataset: removing image inputs results in a noticeable drop in performance. For instance, Qwen2.5 experiences a decline of 0.2 to 0.3 points across BLEU, METEOR, and COMET when visual modality is removed. Similarly, LLaMA3.2 shows decreases of comparable magnitude. This indicates that, in complex or ambiguous scenarios such as those in CoMMuTE, images may serve as useful anchors or sources of semantic disambiguation during decoding. Once visual information is removed, the model loses access to subtle semantic cues that can guide generation, resulting in degraded performance.

Overall, encoder-decoder models are more susceptible to visual noise and struggle with modality alignment, whereas decoder-only models tend to treat visual input as auxiliary guidance that enhances generation. Given the current limitations in aligning image quality and semantics with textual input, removing visual information in encoder-decoder models may act as a form of regularization, whereas for decoder-only models, it may strip away valuable contextual support.

\begin{figure*}[t]
    \centering
    \includegraphics[width=\linewidth]{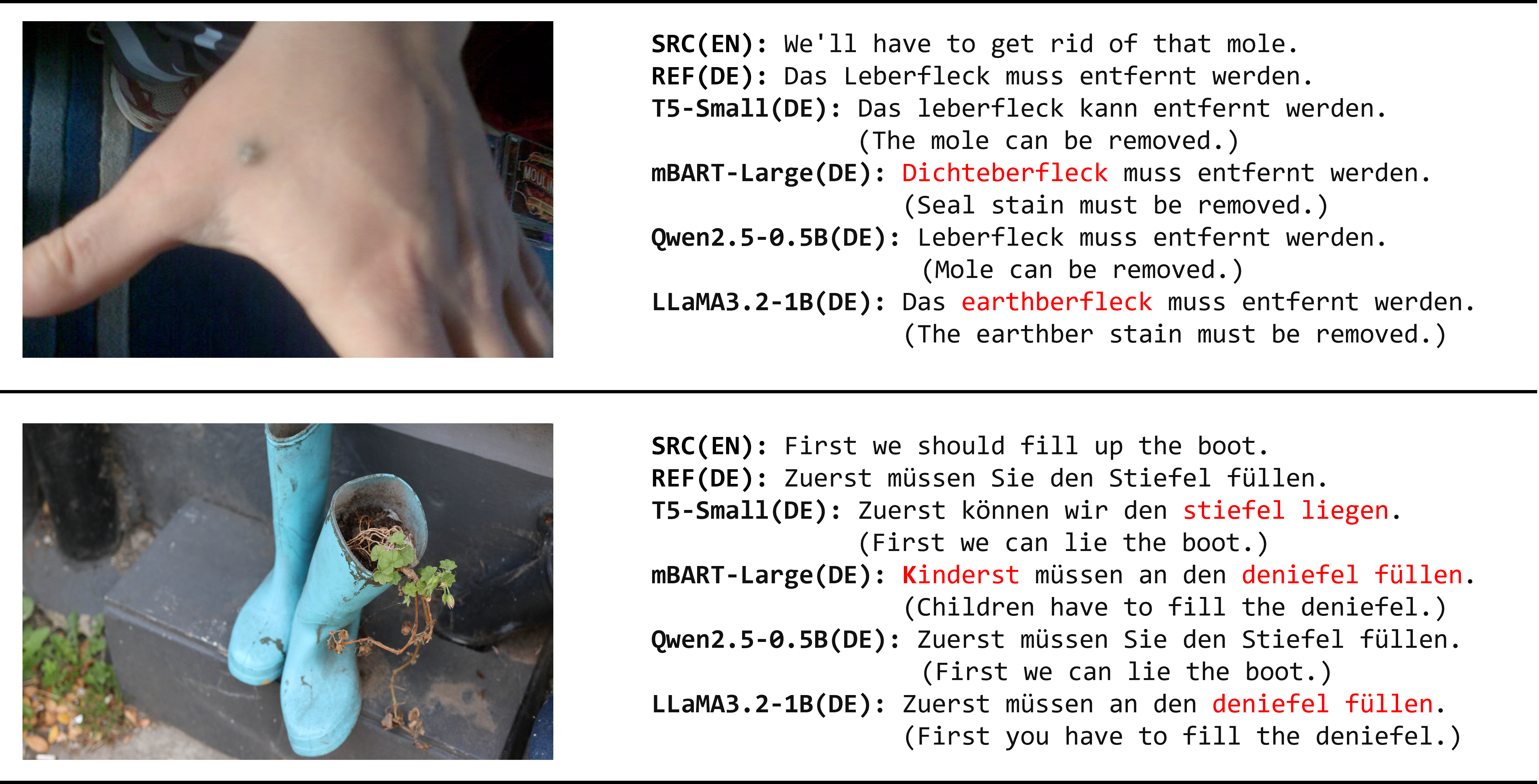}
    \caption{Case study on the CoMMuTE dataset of English-to-German translation direction.}
    \label{fig:case study}
\end{figure*}

\subsection{Sensitivity to Vision-Language Alignment}
MMT systems, when incorporating visual information, are highly sensitive to the semantic alignment between images and text. To systematically evaluate how different pre-trained models respond to the quality of vision-language alignment, we conduct experiments under two conditions: (1) fully aligned image-text pairs and (2) randomly shuffled image-text pairs during the test process, where the image no longer matches the source sentence.

Specifically, we perform this analysis on three standard test sets from the Multi30k dataset as well as the CoMMuTE dataset, which features strong semantic alignment between images and captions. In the shuffled setting, we deliberately disrupt the alignment by randomly pairing each sentence with an unrelated image, potentially introducing misleading visual features and thereby confusing the translation process. To illustrate the impact of alignment quality, Figure \ref{fig:final comparison} presents the average BLEU, METEOR, and COMET scores under both aligned and shuffled conditions across different models and datasets.

As shown in the Figure \ref{fig:final comparison}, all models experience notable performance degradation when image-text alignment is disrupted, confirming that visual features are actively utilized during translation rather than being treated as redundant input. However, the extent of this degradation varies significantly depending on the model architecture.

Firstly, models utilizing pre-trained encoders (e.g., T5 and mBART) exhibit the most severe drops in performance. On the CoMMuTE dataset, for example, COMET scores drop by as much as 5–6 points, while BLEU scores fall by 2–3 points, indicating a strong dependency on visual semantics. This suggests that the stronger the visual encoding capacity, the more sensitive the model is to alignment quality. When image-text pairs are mismatched, these encoders extract irrelevant visual signals, which in turn mislead the translation process.

In contrast, models equipped with pre-trained language decoders (e.g., Qwen and LLaMA) are generally more robust. While they also exhibit slight performance degradation under shuffled conditions, their scores remain relatively stable particularly in terms of BLEU and METEOR. This highlights the advantage of strong language modeling capabilities in pre-trained decoders, which allow the model to partially disregard noisy or misleading visual inputs and maintain coherent translation outputs.

\subsection{Case Study}

As shown in Figure \ref{fig:case study}, we analyze several cases involving semantic ambiguity. The source sentence ``We’ll have to get rid of that mole'' contains the polysemous word ``mole'' which, in this context and given the accompanying image of a hand, clearly refers to a skin blemish. The reference German translation ``Das Leberfleck muss entfernt werden'' correctly captures this meaning using the medically and colloquially appropriate term ``Leberfleck''. Among the four models, Qwen2.5-0.5B successfully reproduces the reference translation, indicating effective disambiguation through multimodal grounding. In contrast, mBART-Large and LLaMA3.2-1B generate hallucinated compound-like terms such as ``Dichteberfleck'' and ``eartherfleck'' likely influenced by visual distractions such as skin tone and lighting.  T5-Small retains the correct noun ``Leberfleck'' but alters the modality of the sentence, translating ``have to'' as ``kann'' (can), which significantly softens the original intent. This example highlights the challenges multimodal systems face when resolving ambiguous terms, especially when the image introduces misleading context or when the model lacks robust domain knowledge.

The second case illustrates how visually salient but semantically irrelevant content can lead to hallucinated outputs. The sentence ``First we should fill up the boot'' is straightforward, and the reference translation ``Zuerst müssen Sie den Stiefel füllen'' reflects a literal and accurate rendering. However, both mBART-Large and LLaMA3.2-1B produce nonsensical tokens such as ``deniefel'' with mBART even hallucinating a subject ``Kinderst'' likely due to the image of child-sized blue boots filled with soil and plants. These outputs reflect a misinterpretation of the visual content, where the models seemingly associate the boots with toys or planters, thereby generating out-of-distribution vocabulary. T5-Small avoids hallucinated words but incorrectly translates ``fill'' as ``liegen'' (to lie), resulting in a semantically erroneous translation. Once again, Qwen2.5-0.5B produces a faithful reproduction of the reference. This case demonstrates that when visual information is prominent yet incongruent with the source text, models may over-rely on vision, leading to severe translation errors. Such errors underscore the importance of effective cross-modal alignment and regularization to mitigate hallucinations driven by visual distractions.

\section{Conclusion}
In this work, we conduct a comprehensive investigation into the role of visual modality in MMT across diverse architectures and pre-trained models. Our findings reveal a clear structural divergence: pre-trained decoders consistently contribute to stable improvements in generation quality, while the benefits of pre-trained encoders are largely contingent on the degree of alignment between visual and textual semantics.

Specifically, decoder-only models, such as Qwen2.5 and LLaMA3.2, exhibit a stronger capacity to leverage visual information as contextual cues during generation, particularly under ambiguous or complex scenarios. These models benefit from treating image features as flexible prompts that complement textual understanding. In contrast, encoder-decoder models, although structurally equipped to jointly encode multimodal inputs, tend to suffer from performance degradation when visual features are misaligned or semantically redundant. The encoded visual noise may interfere with the model’s semantic representation, undermining the potential advantages of multimodal integration.

These results underscore a fundamental challenge in current multimodal systems: effective visual grounding requires not only architectural capacity, but also high-quality cross-modal alignment. Our study suggests that future improvements in vision-language pre-training and data curation, especially in enhancing semantic consistency between modalities, will be critical for fully realizing the potential of MMT.

\bibliography{custom}

\appendix

\end{document}